\documentclass{bmvc2k}

\title{Joint Action Unit localisation and intensity estimation through heatmap regression}

\addauthor{Enrique S\'anchez-Lozano}{Enrique.SanchezLozano@nottingham.ac.uk}{1}
\addauthor{Georgios Tzimiropoulos}{Yorgos.Tzimiropoulos@nottingham.ac.uk}{1}
\addauthor{Michel Valstar}{Michel.Valstar@nottingham.ac.uk}{1}

\addinstitution{
 Computer Vision Lab\\
 University of Nottingham\\
 Nottingham, UK
}
\usepackage{multirow}
\usepackage{graphicx}

\runninghead{Sanchez et al.}{Joint AU localisation and intensity estimation}

\begin{document}

\maketitle

\begin{abstract}
\noindent This paper proposes a supervised learning approach to jointly perform facial Action Unit (AU) localisation and intensity estimation. Contrary to previous works that try to learn an unsupervised representation of the Action Unit regions, we propose to directly and jointly estimate all AU intensities through heatmap regression, along with the location in the face where they cause visible changes. Our approach aims to learn a pixel-wise regression function returning a score per AU, which indicates an AU intensity at a given spatial location. Heatmap regression then generates an image, or channel, per AU, in which each pixel indicates the corresponding AU intensity. To generate the ground-truth heatmaps for a target AU, the facial landmarks are first estimated, and a 2D Gaussian is drawn around the points where the AU is known to cause changes. The amplitude and size of the Gaussian is determined by the intensity of the AU. We show that using a single Hourglass network suffices to attain new state of the art results, demonstrating the effectiveness of such a simple approach. The use of heatmap regression allows learning of a shared representation between AUs without the need to rely on latent representations, as these are implicitly learned from the data. We validate the proposed approach on the BP4D dataset, showing a modest improvement on recent, complex, techniques, as well as robustness against misalignment errors. Code for testing and models will be available to download from \url{https://github.com/ESanchezLozano/Action-Units-Heatmaps}. 
\end{abstract}

\section{Introduction}
\label{sec:intro}
\begin{figure}[t!]
\centering
  \includegraphics[width=0.9\columnwidth]{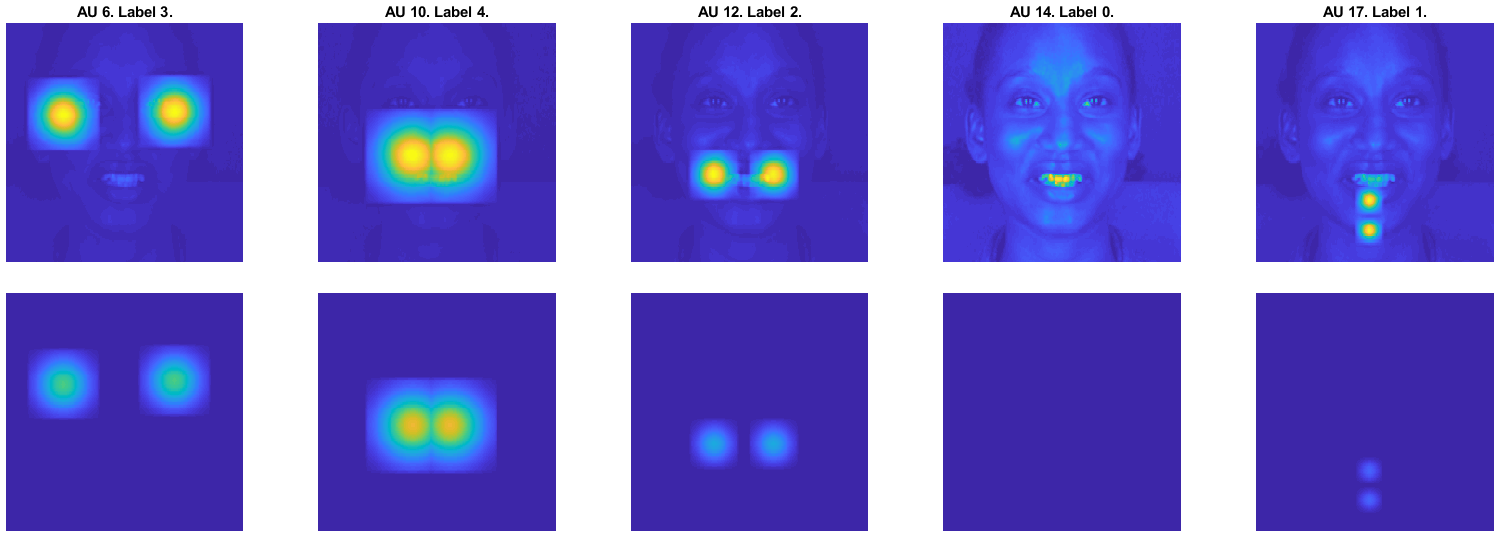}
  \caption{Target heatmaps for a given sample on BP4D~\cite{zhang2014}. The size and peak of the heatmaps are given by the corresponding labels, and are located according to the landmarks defining the AU locations. These heatmaps are concatenated to form the heatmap regression. }
  \label{fig:heatmaps}
\end{figure}
\begin{figure}[t!]
\centering
  \includegraphics[width=\columnwidth]{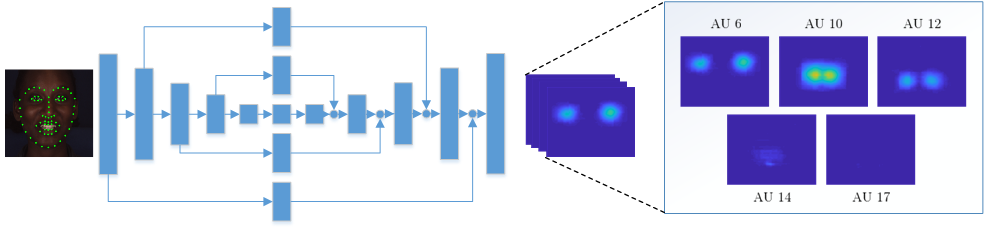}
  \caption{Proposed approach. The facial landmarks are located using a built-in face tracker, and further aligned. The image is then re-scaled to be $256\times256$ pixels, and passed through the Hourglass network, generating the heatmaps corresponding to the Action Units, at the place these occur. The heatmaps are activated according to the predicted intensity. }
  \label{fig:hourglass}
\end{figure}

Automatic facial expression recognition is an active research problem in Computer Vision, which typically involves either the detection of the six prototypical emotions or the Action Units, a set of atomic facial muscle actions from which any expression can be inferred~\cite{ekman02}. It is of wide interest for the face analysis community, with applications in health, entertainment, or human-computer interaction. The Facial Action Coding System (FACS,~\cite{ekman02}), allows a mapping between Action Units and facial expressions, and establishes a non-metric ranking of intensities, spanning values from $0$ to $5$, with $0$ indicating the absence of a specific Action Unit, and $5$ referring to the maximum level of expression. 


Being facial muscle actions, Action Units are naturally correlated with  different parts of the face, and early works in detection and/or intensity estimation used either geometric or local hand-crafted appearance features, or both~\cite{almaev13,valstar2015}. The development of CNN-based methods rapidly took over former appearance descriptors, gaining increasing popularity~\cite{gudi2015}. The advances in facial landmark localisation methods have also permitted to focus on the regions on which an Action Unit is meant to occur, and recent works have taken advantage of this information to exploit a region-based prediction, by which only local CNN representations are used to predict the specific AU intensity level~\cite{jaiswal2016}. It is also known that Action Units rarely occur alone, and most of the times appear combined with others. Based on these facts, some works have attempted to learn a latent representation or graph to model the occurrence of joint Action Units~\cite{sandbach13,tran2017}. This way, the local representations can be combined using the global interaction between AUs. 

All of these advances rely on accurately registering the image to do a proper segmentation, making the models sensitive to alignment errors. Some works have also attempted to perform a joint facial localisation and AU estimation~\cite{wu16}, taking advantage of Multi-task learning to exploit the correlation that exists between the facial expressions and their localisation. However, these models are sensitive to failure when the facial landmarks are incorrectly located. Furthermore, it is worth noting that state of the art methods in Action Unit detection and intensity estimation build on complex deep learning models, which are in practice hard to embed in portable devices, leaving their application to fast GPU-based computers. 

In order to overcome the aforementioned limitations, we propose a simple yet efficient approach to Action Unit intensity estimation, by jointly regressing their localisation and  intensities. In order to do so, rather than learning a model to return an output vector with the AU intensities, we propose to use \textit{heatmap regression}, where different maps activate according to the location and the intensity of a given AU. In particular, during training, a 2D Gaussian is drawn around the locations where the AUs are known to cause changes. The intensity and size of the Gaussians is given by the ground-truth intensity labels. The use of variable-sized Gaussians allows the network to focus on the corresponding intensities, as it is known that higher intensity levels of expressions generally entail a broader appearance variation with respect to the neutral appearance. An example of the heatmaps is depicted in Figure~\ref{fig:heatmaps}. We thus use the Hourglass architecture presented in ~\cite{newell2016}, in the way described in Figure~\ref{fig:hourglass}. We argue that a simple heatmap regression architecture, using a single Hourglass, suffices to attain state of the art results. Our network is shown to deal with facial landmark localisation errors, also presenting a low computational complexity ($\sim 3.5M$ parameters).   

In summary, the contributions of this paper are as follows:
\begin{itemize}
\item We propose a joint AU localisation and intensity estimation network, in which AUs are modeled with heatmaps. 
\item We show that we can handle the heatmap regression problem with a single Hourglass, making the network of low complexity. 
\item We show that our approach is robust to misalignment errors, proving the claim that our network learns to estimate both the localisation and the intensity of Action Units.
\end{itemize}

\section{Related work}
\label{sec:related}
Facial Action Unit intensity estimation is often regarded as either a multi-class or a regression problem. The lack of annotated data, along with the limitation of classic machine learning approaches to deal with unbalanced data, was making early works to model the intensity of facial actions individually~\cite{littlewort11,almaev13}. Recently, the use of structured output ordinal regression~\cite{rudovic12,rudovic13b,rudovic2015,walecki16}, and the refinement of spatial modeling using graphical models~\cite{sandbach13}, shed some improvement in the joint modeling of expressions. In the same line, the use of Multi-task learning has been proposed to model the relation between Action Units~\cite{almaev2015,nicole15}. However, all these approaches need the use of high-level representations of the data, through hand-crafted features.

The use of CNN-based representations has been recently adopted to model facial actions. In~\cite{gudi2015}, a relatively shallow model was introduced, in which a fully connected network was returning the detection and intensity of AUs. Other recent works have attempted to exploit Variational Autoencoders to exploit the latent representations of the data. In~\cite{walecki16}, a Gaussian Process-VAE was introduced to model ordinal representations. In~\cite{walecki17} a Copula-CNN model is proposed jointly with a Conditional Random Field structure to ease the inference. In~\cite{tran2017} a two-layer latent space is learned, in which the second layer is conditioned to the latent representation from the first layer, one being parametric and one being non-parametric. However, despite being accurate, these models require inference at test time, practically slowing down the performance. Besides, these models generally need a feature extraction step: in ~\cite{tran2017} the faces are registered to a reference frame, and given the complexity of the proposed approach, they need to extract a $2000$-D feature vector from a CNN before performing inference in the latent spaces.

In general, the methods described above barely exploit the semantic meaning of Action Units, and they try to learn a global representation from the input image. For these models to succeed, images need to undergo a pre-processing step, consisting of localising a set of facial landmarks and registering the image with respect to a reference frame. Using this information, some recent works have proposed to model Action Units in a region-based approach. In~\cite{jaiswal2016} the face is divided into regions, from which dynamic CNN features are extracted to perform a temporal joint classification. In~\cite{zhao2016} a region layer is introduced to learn the specific region weights, toward making the network focus on specific parts whilst also learning a global representation.

Some works have attempted to model the  spatial location with the Action Unit activations jointly by trying to predict them simultaneously, as both tasks should be highly correlated. In \cite{wu16}, a cascaded regression approach is proposed to learn an iterative update in both the landmarks and the AU detection. In \cite{li2017b} a region-based network is proposed for the task of AU detection, in which a VGG network is first used to extract high-level features, followed by Region of Interest (ROI) networks, the centre of each being located at the spatial location defined by the facial landmarks, and finally topped-up with a fully-connected LSTM network, resulting in a rather complex system. In this paper, rather than approaching a multi-task learning process, we directly propose to estimate the location of the Action Units, along with their intensity, with the latter being the ultimate goal. We show that this approach makes the model less sensitive to errors in  landmark localisation.

\section{Proposed approach}
\label{sec:approach}
The main novelty of our proposed approach resides in the joint localisation of Action Units along with their intensity estimation. This has to be differentiated from works that attempt to jointly estimate the landmarks and detect Action Units. In our framework, we perform a single task learning problem, in which both the localisation and the intensity estimation are encoded in the representation of the heatmaps. The architecture builds on a standard Hourglass network~\cite{newell2016}. Along with the novel learning process, we introduce the use of \textit{label augmentation}, which compensates the lack of annotated data, and increases the robustness of the models against appearance variations. Contrary to previous works on Action Unit intensity estimation, we treat the problem as a regression problem in the continuous range within $0$ and $5$.  
\subsection{Architecture}
The proposed architecture builds on the hourglass network proposed in ~\cite{newell2016}, which has been successfully applied to the task of facial landmark detection~\cite{bulat2016}. The pipeline is depicted in Figure~\ref{fig:hourglass}. The Hourglass is made of residual blocks~\cite{he2016}, downsampling the image to a low resolution and then upsampling to restore the image resolution, using skip connections to aggregate information at different scales. Given an input image, a set of facial landmarks is first located to register the face. The network receives a registered image resized to $256 \times 256$ pixels, and generates a set of $N$ heatmaps, each $64 \times 64$ pixels, where $N$ is the number of Action Units. The $N_j$ map corresponds to an specific Action Unit, which activates at the location the Action Unit is occurring, with a peak and width depending on its intensity.

\subsection{Heatmap generation}
\begin{figure}[t!]
\centering
  \includegraphics[width=0.5\columnwidth]{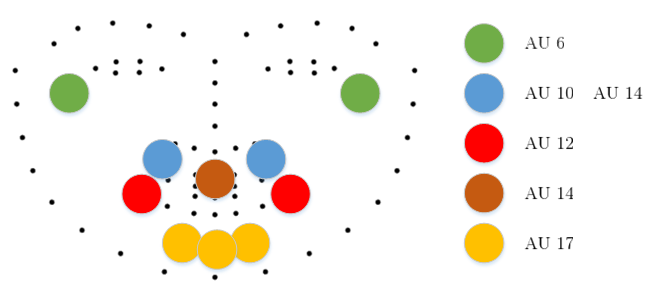} 
  \caption{Gaussian localisation w.r.t the facial landmarks. Each of the circles corresponds to the location of a different AU. Action Units 10 and 14 share two of the points, although each  Gaussian will activate according to their respective labels.}
  \label{fig:truth}
\end{figure}
In this paper, we generate one heatmap per AU. Each heatmap contains two or three Gaussians, as depicted in Figure~\ref{fig:truth}. For a given AU $a$, with labeled intensity $I_a$, a heatmap is generated by drawing a Gaussian at each of its $K$ selected centres, as depicted in Figure~\ref{fig:heatmaps}. Let $(c_x,c_y)$ be a selected centre, the Gaussian $G_k$ is a $64 \times 64$ image, with values at pixel $i,j$ defined by\footnote{Given that input images are of size $256$ and the maps are $64$, the heatmaps are generated by re-scaling the centre accordingly}:
\begin{equation}
\label{heatmap}
G_k(i+c_k,j+c_k) = \begin{cases}
    I_a e^{-\frac{(i^2+j^2)}{2\sigma^2} }      & \quad \text{if } \|i\|, \|j\| \leq 6 I_a \text{ , } \|i\|,\|j\| \in \text{ image bounds }   \\
    0      & \quad \text{otherwise }.
\end{cases}
\end{equation}
The heatmap for Action Unit $a$ is finally given as $H_a(i,j) = \max_k G_k(i,j)$. Thus, the amplitude and the size of the Gaussian is modeled by the corresponding AU label. It is worth noting that the Gaussians are chosen according to visual inspection of where the Action Units cause visible changes in the facial appearance, and these are subject to interpretation or further development. 
\subsection{Loss function}
In heatmap regression, the loss is a per-pixel function between the predicted heatmaps and the target heatmaps, as defined above. The per-pixel loss is defined as the smooth L1 norm (also known as the Huber loss), which is defined as:
\begin{equation}
\mathcal{L}_{i,j} =
        \begin{cases}
        0.5 (x_{i,j} - y_{i,j})^2, & \text{if } |x_{i,j} - y_{i,j}| < 1 \\
        |x_{i,j} - y_{i,j}| - 0.5, & \text{otherwise }
        \end{cases}
\end{equation}
where $x_{i,j}$ is the output generated by the network at pixel $i,j$, and $y_{i,j}$ the corresponding ground-truth. The total loss is computed as the average of the per-pixel loss per Action Unit.

\subsection{Data augmentation}
\label{sec:labaug}
It is a common approach in deep learning to augment the training set by applying a set of random transformations to the input images. In general, augmentation is accomplished by randomly perturbing the pixel values, applying a rigid transformation in scale and in-plane rotation, and flipping the images. In this paper, rather than randomly applying an affine transform, we first perturb the landmarks using a small noise, and then we register the images according to the eye and mouth positions. Thus, a random landmark perturbation not only augments the training data to account for errors in the landmark localisation process, but also implies a random rigid transformation. As it can be seen in Section~\ref{sec:results}, when applying a random landmark perturbation we are implicitly rotating and rescaling the image.

In addition to the input augmentation, we propose to apply a \textit{label perturbation}, by which the labels are also augmented. The motivation behind label augmentation resides in the fact that the manual labeling of intensities is in itself an ambiguous process, and sometimes annotators find it hard to agree on which intensity is exhibited in a given face. Notwithstanding the fact that AU intensity scores are an ordinal scale rather than a metric, we propose to model the intensities as continuous values from $0$ to $5$. We want our network to give a score that is the closest to the ground-truth labels, and thus, during training, labels are randomly perturbed proportionally to their intensities, as follows:
\begin{equation}
I_k \leftarrow 0.2 \, I_k\, |z| \, \mbox{ with } z \sim \mathcal{N}(0,1),
\end{equation}
where $I_k$ is the intensity of the $k$-th AU. This process only affects the heatmap generation, not the loss function during the backpropagation process. 

\subsection{Inference}
Inferring the Action Units' intensities is straightforward in our pipeline. For a given test image, the facial landmarks are first located, and the image is aligned with respect to the reference frame. This image is cropped and rescaled to match the network size, and passed forward the Hourglass. As depicted in Figure~\ref{fig:hourglass}, the network generates a set of heatmaps, each per AU. The intensity estimation consists of simply returning the maximum value of each corresponding heatmap. That is to say, the inference consists of simply performing the $\max$ operator on each of the heatmaps.

\section{Experiments}

\textbf{Set-up:} The experiments are carried out using the Pytorch library for Python~\cite{paszke2017}. The Bottleneck used for the Hourglass has been adapted from the Pytorch \textit{torchvision} library to account for the main Hourglass configuration by which the $3 \times 3$ filters are of $128$ channels rather than the $64$ of the main residual block. As a baseline, we have trained a ResNet-18 using the library as is, modified to account for the joint regression of the $5$ Action Units. 

\textbf{Dataset:} We evaluate our proposed approach on the BP4D \cite{zhang2014,valstar2015} dataset. The \textbf{BP4D} dataset is the corpus of the FERA 2015 intensity sub-challenge \cite{valstar2015}, and it is partitioned into training and test subsets, referred to as BP4D-Train and BP4D-Test, respectively. The training partition consists of $41$ subjects performing $8$ different tasks, eliciting a wide range of expressions. The BP4D-Train subset consists of approximately $140,000$ frames, whereas the B4D-Test is made of $20$ different subjects performing similar tasks, for a total of about $70,000$ frames. For both partitions a subset of $5$ Action Units were annotated with intensity levels. We use a subset of $\sim 80000$ images from the training partition to train our models, and a held-out validation partition of $\sim 40000$ images to validate the trained network. Both partitions capture the distribution of the intensity labels. 

\textbf{Error measures:} The standard error measures used to evaluate AU intensity estimation models are the intra-class correlation (ICC(3,1),~\cite{shrout79}), and the mean squared error (MSE). The ICC is generally the measure used to rank proposed evaluations, and therefore we use the ICC measure to select our models.

\textbf{Pre-processing:} The ground-truth target maps are based on the automatically located facial landmarks. We use the publicly available iCCR code of \cite{sanchez2016, sanchez2017} to extract a set of $66$ facial landmarks from the images. Using these points, faces are registered and aligned with respect to a reference frame. This registration removes scale and rotation. Registered images are then cropped and subsequently scaled to match the input of the network, $256 \times 256$. 

\textbf{Network}: The network is adapted from the Hourglass implementation of the Face Alignment Network (FAN,~\cite{bulat2017}). The Hourglass takes as input a cropped $256\times256$ RGB image, and passes it through a set of residual blocks to bring the resolution down to $64\times 64$, and the number of channels up to $256$. Then, the input $64\times64\times256$ is forwarded to the Hourglass network depicted in Figure~\ref{fig:hourglass}, to generate a set of $5$ heatmaps, each of dimension $64\times 64$. Each of the blocks in the Hourglass correspond to a bottleneck as in \cite{newell2016}. 

\textbf{Training:} The training is done using mini-batch Gradient Descent, using a batch size of $5$ samples. The loss function is defined by the Huber error described in Section~\ref{sec:approach}. We use the RMSprop algorithm~\cite{hinton2012n}, with a learning rate starting from $10^{-3}$. We apply an iterative weighting schedule, by which the error per heatmap (i.e. per AU) is weighted according to the error committed by the network in the previous batch. Each of the weights is defined as the percentage of the error per AU, with a minimum of $0.1$. We use a disjoint validation subset to choose the best model, according to the ICC measure on it.

\textbf{Data augmentation}: We use three types of data augmentation to train our models: \textit{landmark perturbation}, \textit{colour perturbation}, and \textit{label perturbation}. \textit{Landmark perturbation} is used to account for misalignment errors, and consists of adding a small noise to the landmarks before the registration is done, resulting in a displacement of the location where the heatmaps will activate. \textit{Colour perturbation} is used to prevent overfitting, and consists of applying a random perturbation to the RGB channels. Finally, the \textit{label augmentation} is performed as depicted in Section~\ref{sec:labaug}. 

\textbf{Testing}: In order to generate the intensity scores to evaluate the proposed approach, we register the images according to the detected points, to generate the corresponding heatmaps. This process is illustrated in Figure~\ref{fig:hourglass}. Once these are generated, the intensity of each AU is given by the maximum of the corresponding map. We constrain the estimated intensity to lie within the range $0-5$ in the case the maximum is below zero or greater than five. 


\textbf{Models}: As a \textit{baseline}, we have trained a ResNet-18, using the same images, learning rate, and batch size, for the BP4D. The ResNet-18 architecture is modified from the available code to account for joint regression in the 5 target Action Units included in the dataset. This model, which is the smallest ResNet, is made of $\sim 11M$ parameters, almost $4$ times the number of parameters of our network. In order to validate the assumption that the network implicitly learns a shared representation of Action Units, we have trained a set of models each returning a single heatmap, one per AU. We refer to these models as \textit{Single Heatmap}. We compare our method against most recent Deep Learning approaches: the 2-layer latent model of ~\cite{tran2017}, reported to achieve highest ICC score using current deep models, as well as the Deep Structured network of ~\cite{walecki17}, and the GP Variational Autoencoder of ~\cite{eleftheriadis2016}.

\section{Results}
\label{sec:results}
The results are summarised in Table~\ref{au_results2}. It can be seen that our approach outperforms the state of the art in AU intensity estimation, validating the claim that a small network trained to perform AU localisation and intensity estimation suffices to attain state of the art results. It can be seen that the proposed approach clearly outperforms the results given by a deeper model such as the ResNet-18, which attains an ICC score of $0.64$, in contrast to the $0.68$ given by our model. Similarly, it is also shown how the proposed approach learns a shared representation of the AUs, yielding a $3\%$ improvement over training a single hourglass per AU, i.e. a model trained to generate the heatmaps altogether performs better than training a model per AU, being also $5$ times less computationally complex. 

\begin{figure}[h!]
\centering
  \includegraphics[width=0.4\columnwidth]{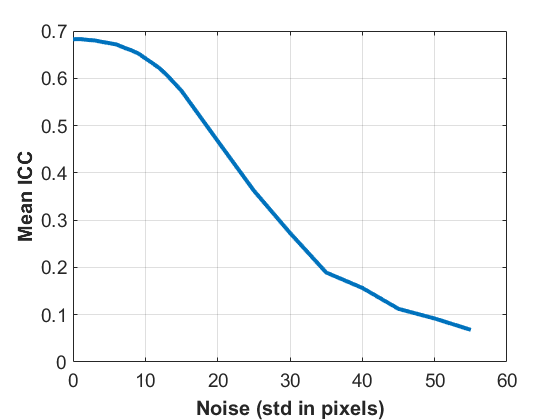} \includegraphics[width=0.4\columnwidth]{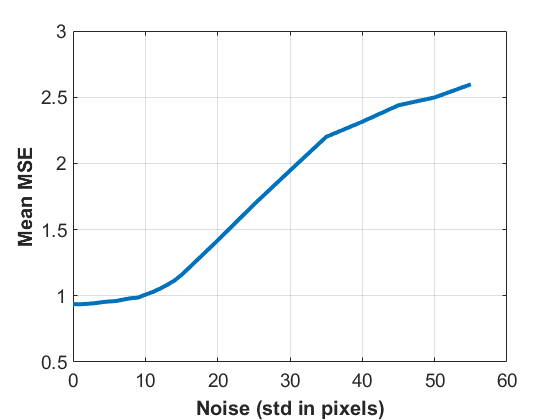}
  \caption{Performance of our network against random noise. The $x$-axis represents the standard deviation of the noise in pixels, whereas the $y$-axis represents the ICC score and the MSE (left and right plots, respectively). }
  \label{fig:noise}
\end{figure}

\begin{figure}[h!]
\centering
  \includegraphics[width=0.9\columnwidth]{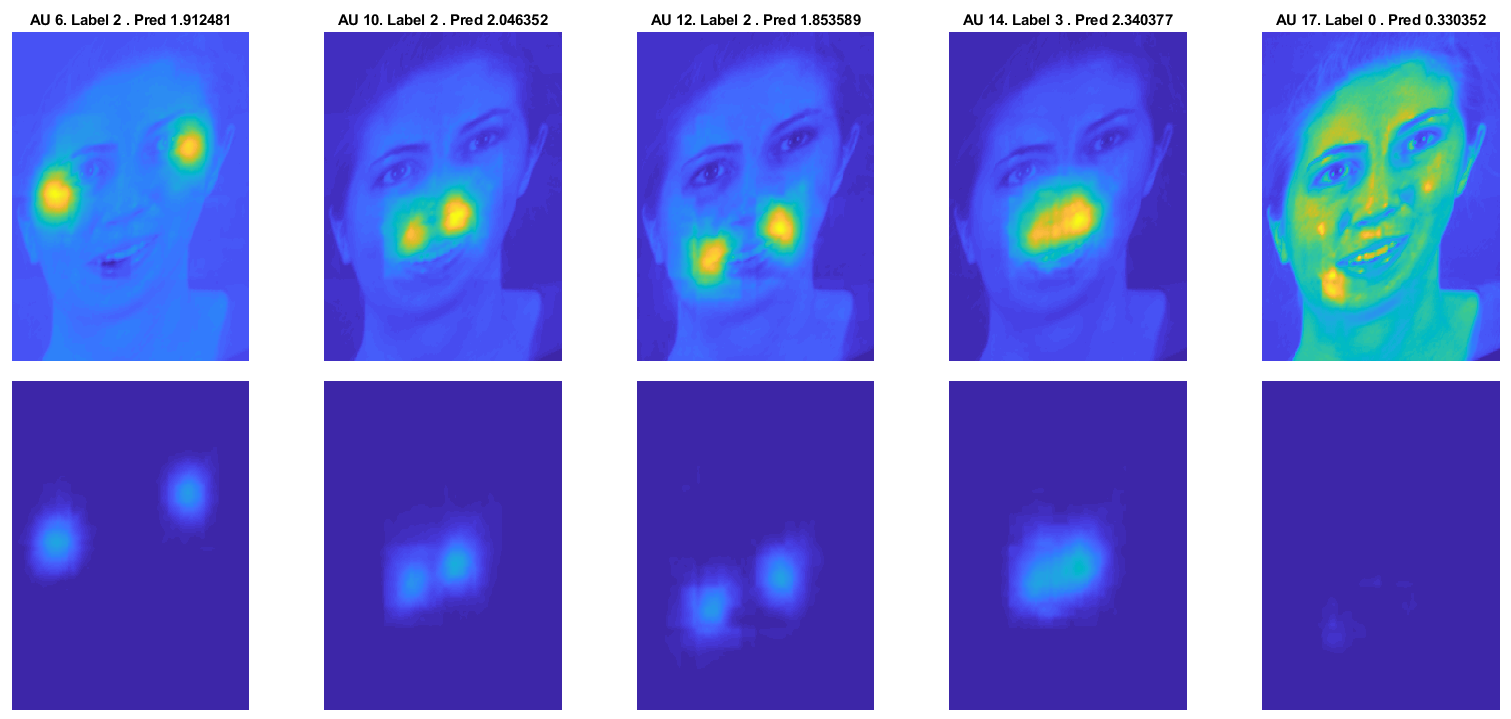}
  \caption{Example of an accurate localisation despite the induced noise. It can be seen how the heatmaps are correctly located, yielding correct predictions.  }
  \label{fig:misa}
\end{figure}

\begin{figure}[h!]
\centering
  \includegraphics[width=0.9\columnwidth]{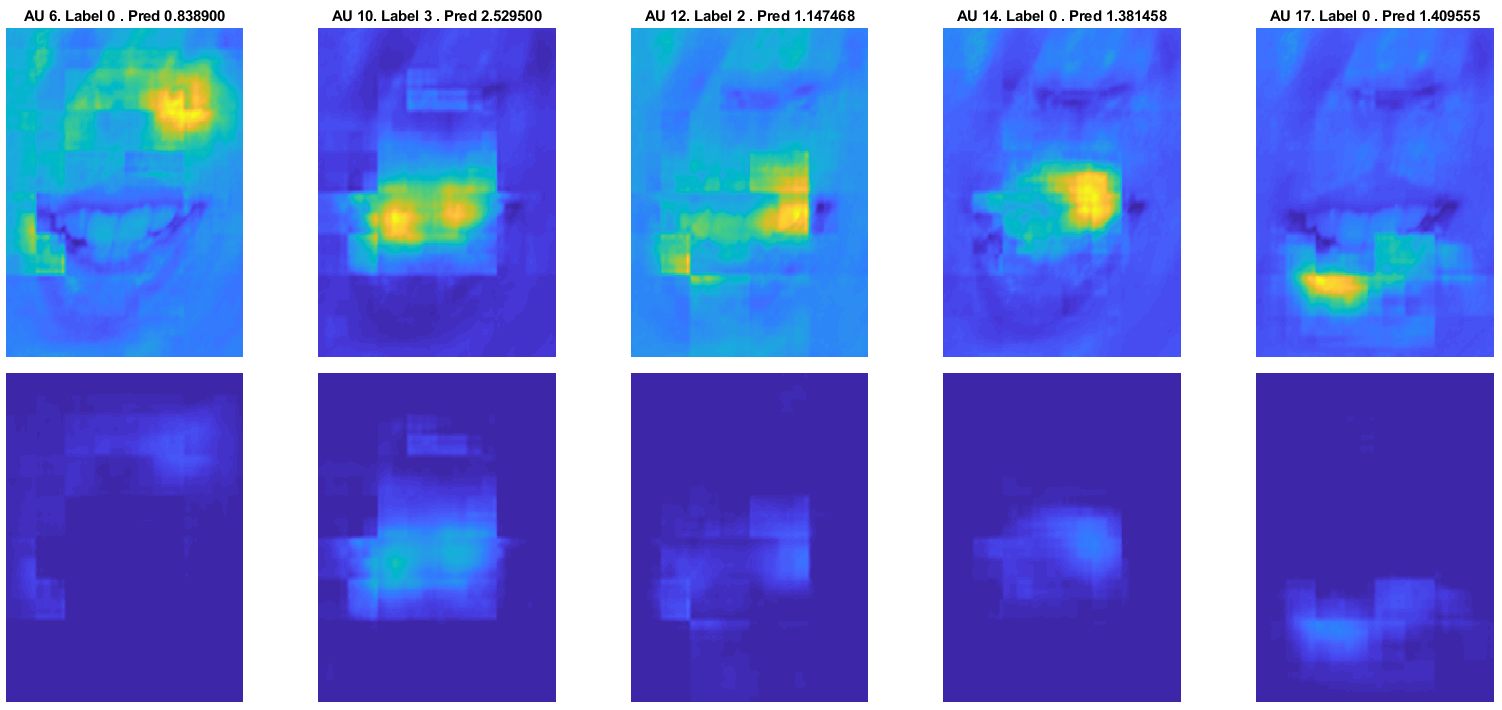}
  \caption{Example of a wrongly predicted image. The noise implies the registration does not allow the image to cover the whole face.  }
  \label{fig:misa2}
\end{figure}

In order to validate the assumption that the network performs the task of AU localisation, we tested our network against random perturbations on the landmarks, that would affect the registration problem, and hence the performance of the network should it hasn't learned to localise the AUs. In particular, we applied a variable Gaussian noise with standard deviations ranging from $0$ to $55$ pixels. The results are reported in Figure~\ref{fig:noise}. It can be seen that our network starts degrading noticeable from a noise $\geq 13$ pixels. With such a noise, the registered images can be heavily distorted, as the noise is applied per landmark. Figure~\ref{fig:misa} shows an example of a good localisation despite the noise, and Figure~\ref{fig:misa2} shows an example of an error caused by a high image distortion. 

\begin{table}[htbp]
\caption{Intensity estimation on BP4D. (*) Indicates results taken from the reference. Bold numbers indicate best performance. $\dagger$ indicates results as reported in \cite{tran2017}}
\label{au_results2}
\begin{center}
\begin{tabular}{c | c | c | c | c | c | c | c |}
  \hline
		& AU	& 6 & 10 & 12 & 14 & 17 & Avg.\\
	\hline
	\multirow{6}{*}{\rotatebox{90}{ICC}} & Our & {\bf 0.79} & {\bf 0.80} & 0.86 & {\bf 0.54} & 0.43 & {\bf 0.68} \\ 
    & Single Heatmap &  0.78 & 0.79  &  0.84 & 0.36 & 0.49 & 0.65 \\
    & ResNet18 & 0.71 &0.76 & 0.84 & 0.43 & 0.44 & 0.64 \\
    & 2DC \cite{tran2017}* & 0.76 & 0.71 & 0.85 & 0.45 & {\bf 0.53} & 0.66 \\ 
    & CCNN-IT \cite{walecki17}* & 0.75 & 0.69 & 0.86 & 0.40 & 0.45 & 0.63 \\
    & VGP-AE \cite{eleftheriadis2016}* & 0.75 & 0.66 & {\bf 0.88} & 0.47 & 0.49 & 0.65 \\
    \hline
	\multirow{6}{*}{\rotatebox{90}{MSE}} & Our	& 0.77 & 0.92 & {\bf 0.65} & 1.57 & {\bf 0.77} & {\bf 0.94} \\
    & Single Heatmap &  0.89 & 1.04 & 0.82 & 2.24 & 0.78 & 1.15 \\
    & ResNet18 & 0.98 & {\bf 0.90} & 0.69 & 1.88 & 0.95 & 1.08 \\
    & 2DC \cite{tran2017}* & {\bf 0.75} & 1.02 & 0.66 & {\bf 1.44} & 0.88 & 0.95 \\
    & CCNN-IT \cite{walecki17} $\dagger$ & 1.23 & 1.69 & 0.98 & 2.72 & 1.17 & 1.57 \\
    & VGP-AE \cite{eleftheriadis2016}* & 0.82 & 1.28 & 0.70 & {\bf 1.43} & {\bf 0.77} & 1.00 \\
	\hline
\end{tabular}
\end{center}
\end{table}


\section{Conclusions}
In this paper, we have presented a simple yet efficient method for facial Action Unit intensity estimation, through jointly localising the Action Units along with their intensity estimation. This problem is tackled using heatmap regression, where a single Hourglass suffices to attain state of the art results. We show that our approach shows certain stability against alignment errors, thus validating the assumption that the network is capable of localising where the AUs occur. In addition, we show that the learned model is capable of handling joint AU prediction, improving over training individual networks. Considering that the chosen ground-truth heatmaps arise from a mere visual inspection, we will further explore which combination might result in more accurate representations. The models and the code will be available to download from \url{https://github.com/ESanchezLozano/Action-Units-Heatmaps}.

\section{Acknowledgments}
This research was funded by the NIHR Nottingham Biomedical Research Centre. The views expressed are those of the authors and not necessarily those of the NHS, the NIHR or the Department of Health.
\bibliography{egbib}
\end{document}